%% file: main.tex
\documentclass[runningheads]{llncs}

\usepackage[utf8]{inputenc} 
\usepackage[T1]{fontenc}    
\usepackage{times}
\usepackage{latexsym}

\usepackage{microtype}

\usepackage{xspace}

\usepackage{microtype}
\usepackage{graphicx}
\usepackage{booktabs}

\usepackage{multirow}
\usepackage{caption}
\usepackage{subcaption}
\usepackage{adjustbox}
\usepackage{amsfonts}

\newcommand{\citep}[1]{\cite{#1}}

\begin{document}
\title{On the Granularity of Explanations in Model Agnostic NLP Interpretability}
\titlerunning{Granularity of Explanations in Model Agnostic NLP Interpretability}
%
\author{Yves Rychener\inst{1} \and
Xavier Renard\inst{2} \and Djamé  Seddah \inst{3} \and Pascal Frossard\inst{1} \and Marcin Detyniecki\inst{2,4,5}}
\authorrunning{Yves Rychener et al.}
%
\institute{EPFL, Lausanne, Switzerland\and
AXA, Paris, France\and
Inria, Paris, France\and
Sorbonne Université, Paris, France \and
Polish Academy of Science, Warsaw, Poland}

\maketitle              

\begin{abstract}
Current methods for Black-Box NLP interpretability, like LIME or SHAP, are based on altering the text to interpret by removing words and modeling the Black-Box response. In this paper, we outline limitations of this approach when using complex BERT-based classifiers: The word-based sampling produces texts that are out-of-distribution for the classifier and further gives rise to a high-dimensional search space, which can't be sufficiently explored when time or computation power is limited. Both of these challenges can be addressed by using segments as elementary building blocks for NLP interpretability. As illustration, we show that the simple choice of sentences greatly improves on both of these challenges. As a consequence, the resulting explainer attains much better fidelity on a benchmark classification task.
\end{abstract}

\section{Introduction and Related Work}
\label{sec:intro}
\input{sections/introduction}

\section{Limits of Word-Based Black-Box Interpretability}
\label{sec:word-problems}
\input{sections/word_problems}

\section{Sentence-Based Interpretability}
\label{sec:gutek}
\input{sections/gutek}

\section{Fidelity Experiment}
\label{sec:experiments}
\input{sections/experiments}

\section{Discussion}
\label{sec:discussion}
\input{sections/discussion}

\section{Conclusion}
\label{sec:Conclusion}
\input{sections/conclusion}
\bibliographystyle{splncs04}
\bibliography{acl2021}

\newpage
\clearpage
\appendix
\input{sections/appendix}

\end{document}

%% file: sections/introduction.tex


Interpretability of Natural Language Processing (NLP) models can be addressed by developing inherently interpretable classification models \citep{lei2016rationalizing,chang2019game,jain2020learning} or with Post-Hoc interpretability that can be applied to already trained models. With the latter, neural network architectures can be interpreted by white box approaches, which need access to model internals like gradients and activations~\citep{arras2016explaining,dimopoulos1995use}. Patterns in attention layers are also used, but the validity of this practice has been under heavy discussion, see \citep{bibal2022attention} for an overview of recent literature in this domain. However, when model access is not possible or preprocessing methods hinder gradient flow, a Black-Box approach without model access is more suitable.
Models like LIME~\citep{ribeiro2016lime} and SHAP~\citep{lundberg2017shap} are examples of Black-Box interpreters which can be applied to texts. They create an interpretation for a text sample, called \textit{local} interpretation. To this end, a dataset of similar texts, called the neighborhood, is sampled by repeatedly removing words from the original text and observing the change in output. The local behaviour of the model is then approximated using a regression on the presence of words, whose weights are interpreted as local effects of the word presence on the prediction. While LIME and SHAP perform the sampling of the neighborhood directly in the text domain, other approaches use for example auto-encoders to generate neighboring texts~\citep{lampridis2020explaining}. While such approaches are promising, their performance heavily depends on the performance of the text generation model. Since in practice, resampling in the text domain is still the most prevelant, we will consider this approach in this work.

We explore the limits of the approach of using words when it comes to complex language models like BERT~\citep{devlin2019bert}.
In concurrent work, Zafar et al. \cite{zafar2021more} also investigate if sentences are more suitable for NLP interpretability. They find that sentence interpretations are more robust than word based interpretations and lead to lower variability when using approximation techniques. We hypothesize that these two results may be direct consequences of the results in Sections~\ref{sec:word-problems:dist-shift} and \ref{sec:word-problems:complexity} respectively.
Our work can thus be seen as complementary to \cite{zafar2021more}, as it confirms the results independently and gives interpretation for the source of the better performance of sentence-based methods.
Our main contributions are the identification of the granularity (words/sentences/paragraphs) as a crucial, often overlooked hyper-parameter in black-box NLP interpretability. In addition to displaying the problems arising from this negligence, we show that an interpreter using sentences as elementary units is able to greatly address the identified problems. Finally, we achieve substantially higher performance in the benchmark problem used for assessing fidelity to the underlying classifier. With this work, we hope to spark a discussion in the literature about the importance of granularity for NLP interpretability.






%% file: sections/word_problems.tex
While removing words to interpret a model is suitable for Bag-Of-Words (BOW) models without n-grams, the use of models like BERT~\citep{devlin2019bert}, which try to model word interactions using the attention mechanism, warrants a discussion if this is the appropriate sampling mechanism for such models: Removing random words from a text can make it unreadable for humans, since key interactions, like verb-subject, are broken. Is this also observed with BERT? 
What are other consequences of word-based sampling?
We compare the commonly used word based sampling to sentence-based sampling. We argue it is a more natural choice for interpretability, since sentences represent syntactically closed units and can greatly reduce the dimensionality of the neighborhood to explore.

\subsection{Distributional Shift}
\label{sec:word-problems:dist-shift}
Sampling the neighborhood is done by altering the text. The sampling mechanism thus has an effect on the embedding of the altered text.
For neural networks, it is well studied that the Out-Of-Distribution (OOD, different distribution than training distribution) performance can be significantly worse than In-Distribution (ID, same distribution as training data) performance~\cite{nguyen2015ood,amodei2016ood,hendrycks2016ood,liang2017ood,moosavi2017ood}, with sometimes dramatic errors known as adversarial attacks. In order for the explanation, which is based on the altered texts, to be truthful, downstream classifier accuracy must be maintained for those altered texts. This can only be guaranteed if texts remain in-distribution after alteration, which we will show is not the case with word-based sampling.



Consider a simple example: Assume a perfect classifier which is able to correctly identify the sentiment of any natural text. However, if the text does not contain a verb, it predicts the opposite sentiment. Consider now the text "The food was nice.". Any text alteration method, which removes the verb, produces an adversarial text for which the model makes an incorrect prediction. An explanation based on such an alteration method does not reflect the behaviour of the classifier on natural (ID) text.
Inspired by \cite{lee2018simple} where hidden activations were used to detect OOD samples for images, we use the text embedding produced by language models (\texttt{[CLS]}-Token) to detect distributional shift in two experiments. This is because in many approaches, the \texttt{[CLS]}-Token is used as an input for downstream models, which may receive an OOD input.


\subsubsection{Visualizing Distributional Shift}
\label{sec:word-problems:dist-shift:1}

\begin{figure}
    \centering
    \includegraphics[width=\linewidth]{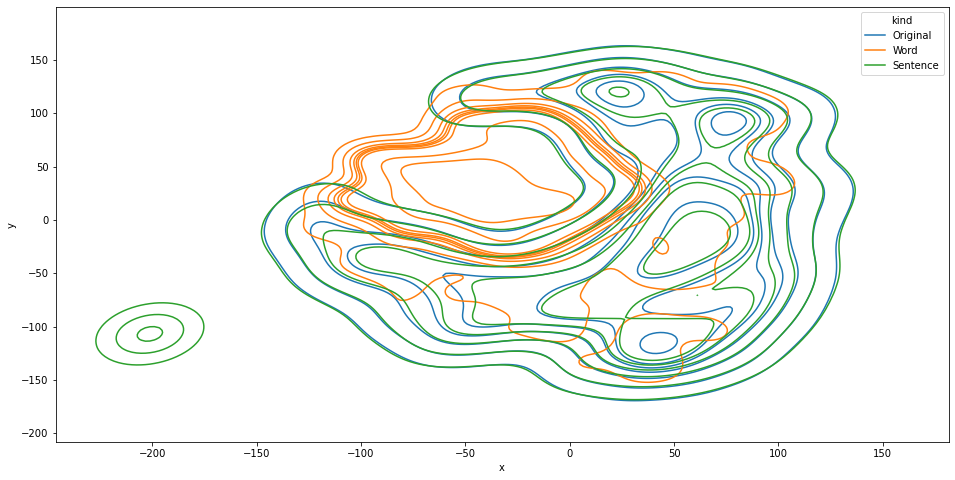}
    \caption{t-SNE of Distributional Shift with 10,000 samples. $W_1(words)=8.6$, $W_1(sentence)=4.1$}
    \label{fig:ood}
\end{figure}
In the first experiment, we compare the distribution of the embeddings of the original text, after removing a random sentence and after randomly removing the same number of words. We compute the embeddings for 10,000 randomly selected Wikipedia snippets from the SQuAD dataset~\citep{rajpurkar2016squad} using BERT~\citep{devlin2019bert}. On a t-SNE visualisation (Figure~\ref{fig:ood}) of the distributions of the embeddings (original text, sentence removed, words removed) one can observe that the distribution obtained by removing randomly selected words (orange) is significantly different from the original one (blue), while no big difference is observed with removing sentences (green). 
To quantify this effect, we consider the \textit{Wasserstein Distance}. Given two distributions $\mathbb{P}$ and $\mathbb{Q}$, it is defined as
$$
W_1(\mathbb{P}, \mathbb{Q}) = \min_{\pi\in\Pi(\mathbb{P}, \mathbb{Q})}\mathbb{E}_{(x,y)\sim\pi}[\|x-y\|],
$$
where $\Pi(\mathbb{P}, \mathbb{Q})$ is the set of all couplings between $\mathbb{P}$ and $\mathbb{Q}$. 
The Wasserstein Distance or "earth mover distance" measures the minimum cost (probability mass multiplied by distance moved) to turn one probability distribution into another.
We now consider by $\mathbb{Q}$ the empirical distribution of the embeddings of original text, $\mathbb{P}_{s}$ the empirical distribution of the embeddings of texts with a sentence removed and $\mathbb{P}_{w}$ the empirical distribution of the embeddings of texts with words removed. We obtain $W_1(\mathbb{P}_w, \mathbb{Q})=8.6$ and $W_1(\mathbb{P}_s, \mathbb{Q})=4.1$, which confirms that texts with sentences removed are closer to standard text than texts with words removed. Since the classifier is trained on normal texts, its accuracy on the texts obtained by word-sampling, as used by current state of the art model-agnostic interpretability methods, is questionable, since they are OOD. However, sentence sampling produces ID texts, for which normal accuracy can be expected.


\subsubsection{Evaluating Distributional Shift with Classifier Accuracy}
\label{sec:word-problems:dist-shift:2}
\begin{figure}
    \centering
    \includegraphics[width=\linewidth, trim= 2cm 1cm 3cm 1.5cm]{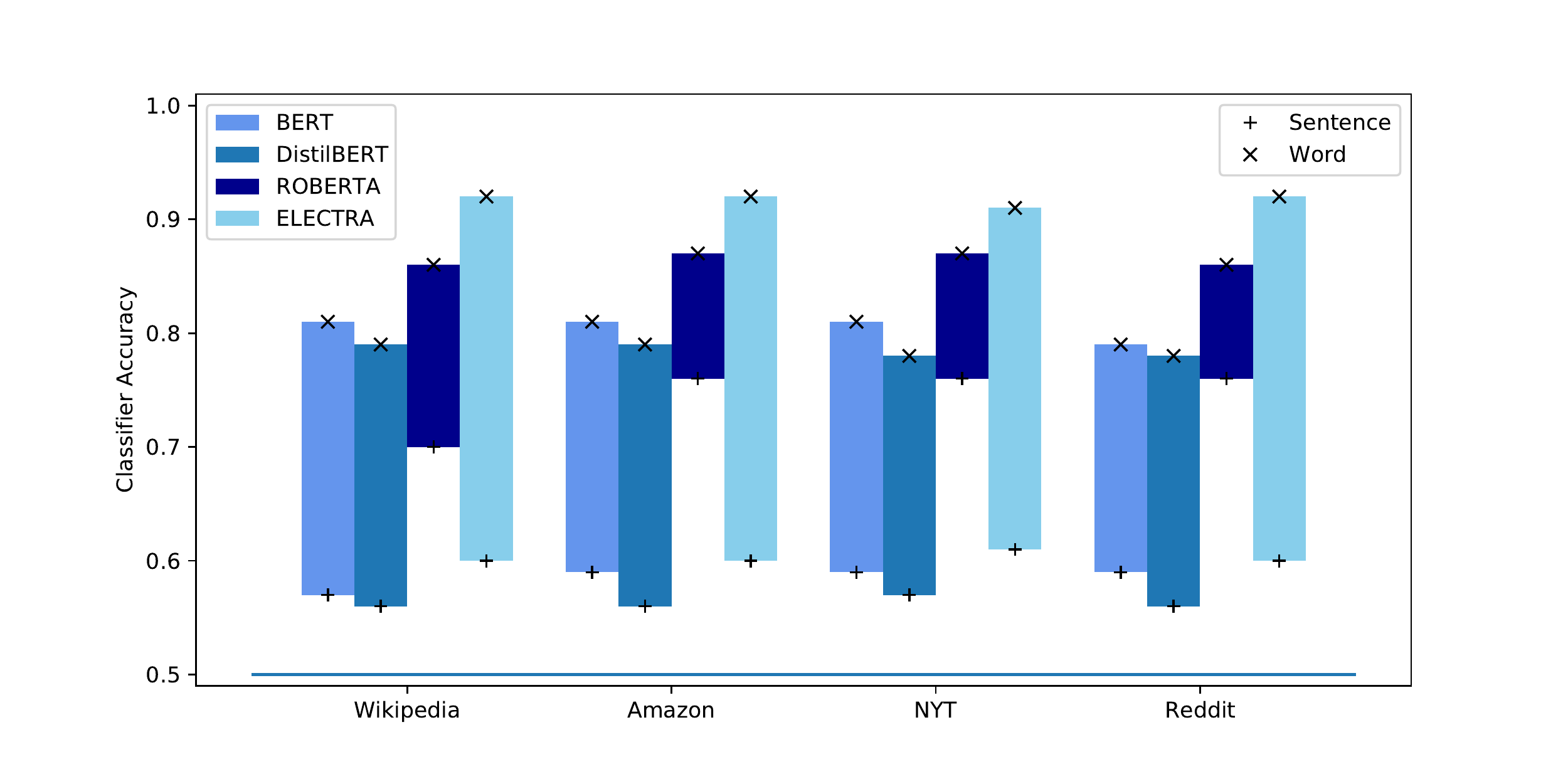}
        \caption{Comparing Distributional Shift Classifier Accuracy}
        \label{fig:ood:classifier}
\end{figure}
One may wonder if the distributional shift observed in the previous experiment is only because a relatively high number of words was removed, reflecting a strong alteration of the text. We perform a second experiment, by framing the detection of distributional shift as a classification problem: The classifier is given text embeddings and tasked with distinguishing between altered and unaltered texts.
We compare the embeddings of the original texts, with 5 words removed and with 1 sentence removed. In order to further study if the distributional shift effect is present across different pretraining schemes and prevails after distillation, we use a range of language models other than BERT~\citep{devlin2019bert}, namely DistilBERT~\citep{sanh2019distilbert}, ROBERTA~\citep{liu2019roberta} and ELECTRA~\citep{clark2020electra}, where DistilBERT is a distilled version of BERT, while ROBERTA and ELECTRA use different pretraining tasks, notably loosing next sentence prediction. We employ a variety of different text domains by using context from SQuAD 2.0~\citep{rajpurkar2018squad} and SQuADShifts~\citep{miller2020squadshifts}. While SQuAD 2.0 contains texts from Wikipedia, SQuADShifts contains texts from other domains, which are often encountered in practice. These include \textit{user generated text} (from Amazon reviews and Reddit comments) and \textit{newspaper articles} (New York Times).
For each binary classifications (Original-Word and Original-Sentence for all datasets and transformer models), we train a Random Forest Classifier on the embeddings and observe its performance on a randomly selected, held-out test set.
The results are given in Figure~\ref{fig:ood:classifier}. Since the binary classifications are balanced, random predictions would yield a classification accuracy of $0.5$. We observe that for all datasets and Language Models, the classification accuracy for sentence-removal is much lower compared to word-removal, almost down to random prediction. This suggests that the distributional shift with sentence-removal is much lower, confirming the results from Section~\ref{sec:word-problems:dist-shift:1}.
The fact that the result is not only observed on the Wikipedia subset, but also Amazon, New York Times and Reddit, suggests that distributional shift is a problem across text domains and transformer-based Language Models. Further, using sentences seems to successfully address the issue for most language models except ROBERTA, where the altered text seems to still be OOD, although an improvement can be observed, indicated by the lower accuracy. While this behaviour of the different language models is an interesting property, we leave its analysis for further works. For the arguments presented here, it suffices to note that sentence based interpretability shows preferable distributional properties with reduced distributional shift.

\subsection{Computational Complexity}
\label{sec:word-problems:complexity}
Since language models often require substantial computation power, even in inference, computational complexity is another issue with word based methods. 
We can view the sampling from the neighborhood as sampling binary vectors, encoding the presence/absence of words or sentences, where the number of possible choices, i.e. the size of the neighborhood is exponential in the number of words/sentences. 
Taking the SQuAD 2.0~\citep{rajpurkar2018squad} dataset for illustration, the texts contain on average 137.7 words in 5.1 sentences. The number of elements in the neighborhood are thus $2^{137.7}=2.8*10^{41}$ for word-based alteration and $2^{5.1}=34.3$ for sentence-based alteration. Since in practical applications, computation time is often constrained, only a limited number of samples from the neighborhood can be evaluated. Since the neighborhood resulting from sentence-based alteration is much smaller, a higher portion of it can be explored. If for example time permits only exploring 20 samples, then $58\%$ of the sentence-based neighborhood can be explored. However, of the word based neighborhood, less than $10^{-39}\%$ can be explored. This results in a better estimation of the model's decision surface with sentence-based methods when computation power is limited.

%% file: sections/gutek.tex
To explain a sample, standard post-hoc model-agnostic interpretability approaches create a dataset of the local neighborhood by repeatedly perturbing parts of the input. The created dataset is then used to train an interpretable surrogate model, for example a linear regression, on the model predictions.

Based on the insights from Section~\ref{sec:word-problems}, we propose to use sentences as atomic units for explanations.
In addition to sentence-based alteration, we use a different methodology to select which parts to alter. For tabular data, \cite{laugel2018locality} conclude that defining locality is a crucial issue for local Black-Box interpretability. We hypothesize that the same holds for text classification: Texts should be sampled such that small changes are more frequent than large changes. This is why we use the most \textit{local} neighborhood possible: we enumerate the alterations with the fewest sentences removed. Since the dataset of the neighborhood is well localized, using a weighted regression like in LIME or SHAP is not necessary.

We propose the GUTEK\footnote{GUTEK, ``Gutenberg'' in Polish, for \textbf{\underline{G}}enerating \textbf{\underline{U}}nderstandable \textbf{\underline{T}}ext \textbf{\underline{E}}xplanations based on \textbf{\underline{K}}ey segments.} approach in three steps:
We first split the text into sentences (\textbf{Segmentation}). We then repeatedly remove some sentences in order to create a dataset reflecting the local neighborhood of the sample to explain (\textbf{Local Sampling}). This dataset is then used to fit a linear regression on the presence/absence of sentences (\textbf{Surrogate Training}). The weights of the regression can be interpreted as the local effect of the presence of sentences on the prediction.

%% file: sections/experiments.tex
In Section~\ref{sec:word-problems} we point out the main reasons for proposing sentence-based interpretability: computational complexity and distributional shift. While we give theoretical arguments why these are important drawbacks of word-based methods, we ultimately want to give \textit{better} explanations. Defining what is a \textit{good} explanation is still an open question in interpretability research, but we identify \textbf{fidelity} as a desirable property. This means that the given explanation well reflects the reasoning of the underlying classifier.

In order to assess if GUTEK correctly explains the classifier's reasoning, we test if it is able to detect which parts of the text were important for the prediction. We use the  QUACKIE~\citep{rychener2020} benchmark.
QUACKIE aims to address the human bias in the ground-truth generation for NLP interpretability tasks. This is done by, instead of human annotating ground truth labels for existing classification tasks, constructing a specific classification task for which the ground-truth labels arise directly from the underlying dataset. That is, for a given question-context pair in Question-Answering datasets, the classification models are tasked with determining if the question can be answered with the context. The sentence containing the answer in the context is then used as ground-truth interpretability label. QUACKIE comprises three performance metrics, namely IOU, calculated as the intersection-over union in terms of sentences and measuring how well the ground truth has been found, HPD, computing inverse rank of the ground truth sentence and SNR, computing the square of the score of the important sentence divided by the variance of the scores of unimportant sentences.

We compare our approach to LIME with \textit{sum} aggregation of token scores for each sentence, which represents the current best-performing Black-Box method in the benchmark in the primary metrics IoU and HPD, representing performance of \textit{correctly identifying the important sentence} and \textit{highly ranking the important sentence} respectively.
We report the results for the SQuAD 2.0 dataset in Table~\ref{table:res_squat}, results from other domains, such as Reddit posts or New York times Articles, are given in Appendix~\ref{appendix:quackie results} and show the same behaviour. We outperform the previous method by a substantial margin in both IoU and HPD for both classifiers. Notably in IoU, our approaches scores are more than double LIME's scores with the same number of samples, which implies that we find the most important sentence twice as often as the word-based approach. When allowing LIME 10 times as many samples as our approach (100 samples vs. 10 samples for GUTEK) it gets closer to our performance in IoU and HPD without matching it.\footnote{The scores are also better than the ones obtained for LIME on a random subset of samples using a neighborhood of 1000 samples.} 
Obviously, drawing 10 times as many samples also results in roughly a 10 fold increase in required computation power and thus a roughly 10 fold increase in computation time. Using 100 samples with the sentence-based approach results in a minor improvement of about 3 percentage points in the primary metrics IoU and HPD, suggesting that the neighborhood is already sufficiently well explored with 10 samples. In the SNR metric, measuring \textit{how much higher the score for the important sentence is compared to the unimportant ones}, LIME is performing better than GUTEK, possibly due to the use of LASSO regression, which was pointed out by the benchmark authors as a possible attack to improve the SNR score.
Overall, the explanations from the sentence-based approach thus better represent the model's reasoning.
\begin{table}[t]
\begin{center}
\begin{small}
\begin{sc}
\resizebox{0.6\linewidth}{!}{\begin{tabular}{l|rrr|rrr}
\toprule
& \multicolumn{3}{c|}{Classif} & \multicolumn{3}{c}{QA}\\
Method & IoU & HPD & SNR & IoU & HPD & SNR\\
\midrule
GUTEK 10 &\textbf{88.55}&\textbf{90.75}&\textbf{39.48}&\textbf{90.53}&\textbf{92.37}&37.37\\
LIME 10 &37.70&50.29&39.23&38.47&50.83&38.20\\
LIME 100 &58.04&66.50&39.30&69.90&75.98&\textbf{40.91}\\
\bottomrule
\end{tabular}}
\end{sc}
\end{small}
\end{center}
\caption{Results on QUACKIE (SQuAD)}
\label{table:res_squat}
\vskip -0.1in
\end{table}
We hypothesize that the improvement in fidelity is due to the reduced distributional shift (Section~\ref{sec:word-problems:dist-shift}) and much reduced search space (Section~\ref{sec:word-problems:complexity}). This is in line with the observation that LIME is able to improve its performance when using more samples.

%% file: sections/discussion.tex
We have shown that word-based sampling mechanisms which alter the text by removing words create a distributional shift in the input texts. This may lead to OOD inputs to the underlying model when the neighbourhood is explored. We further showed that the neighborhood created by word-based methods is very big and can not be well explored with a limited number of samples. While using an iterative approach, first finding the important sentences in a text, then the important words in the important sentences would address the second problem, the first would prevail.

While in this work, we used sentences as elementary building blocks for NLP interpretability, this is not the best choice in all applications. For example, in short texts like tweets, where less interdependence between words is present, word-based approaches may be preferable. Similarly, for very long texts where there is a strong interdependence between sentences, even bigger segments, such as paragraphs, may be used. Finally, also parts of sentences may be used. However, this raises the problem of text segmentation, which is beyond the scope of this paper.

\begin{figure*}
    \centering
    \begin{subfigure}[c]{0.6\textwidth}
    \begin{center}
    \includegraphics[width=\linewidth, trim=5cm 0 24cm 0, clip]{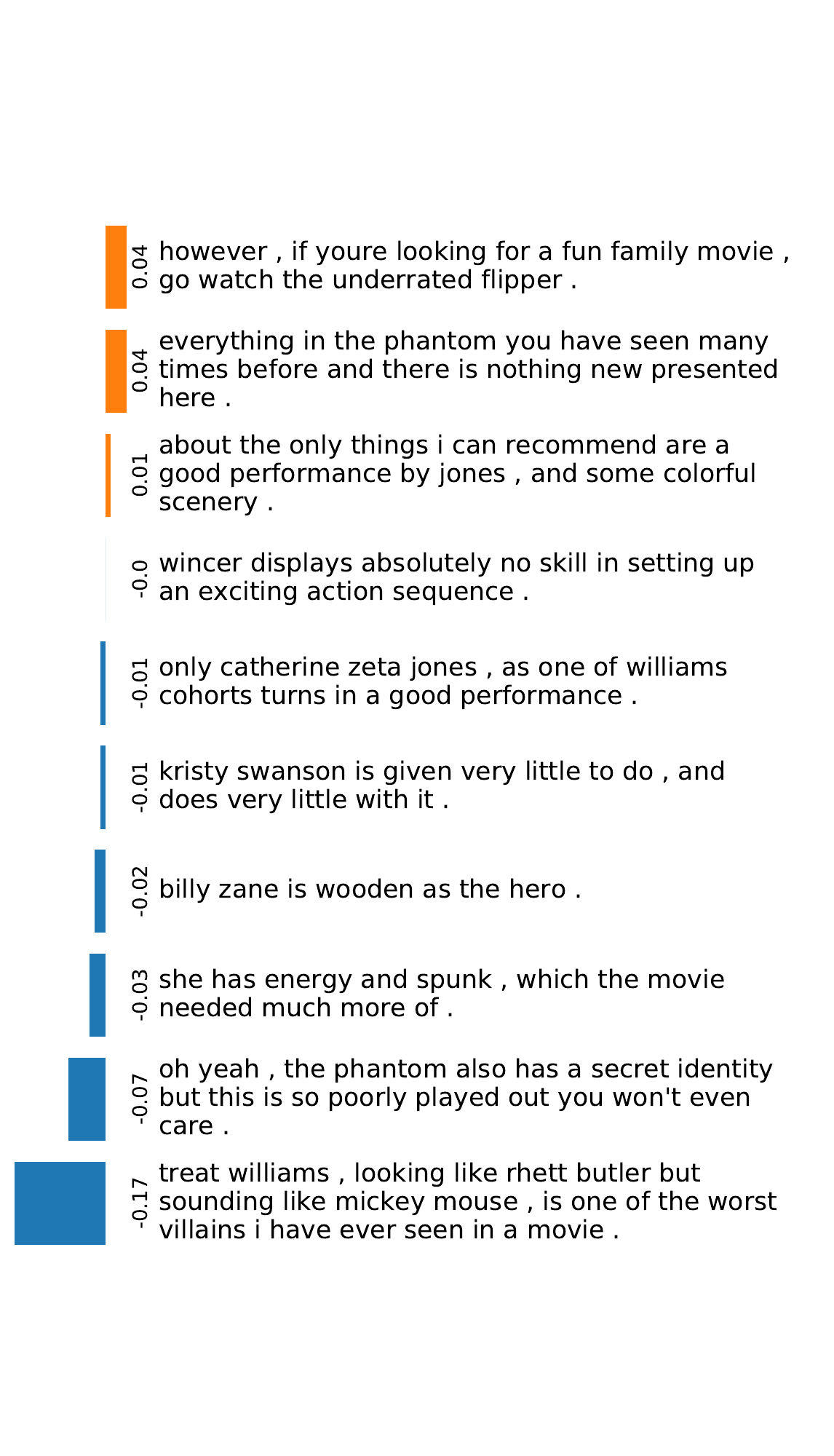}
    \end{center}
    \end{subfigure} 
    \begin{subfigure}[c]{0.3\textwidth}
    \begin{center}
        \includegraphics[width=\linewidth, trim=13cm 20cm 27cm 0cm, clip]{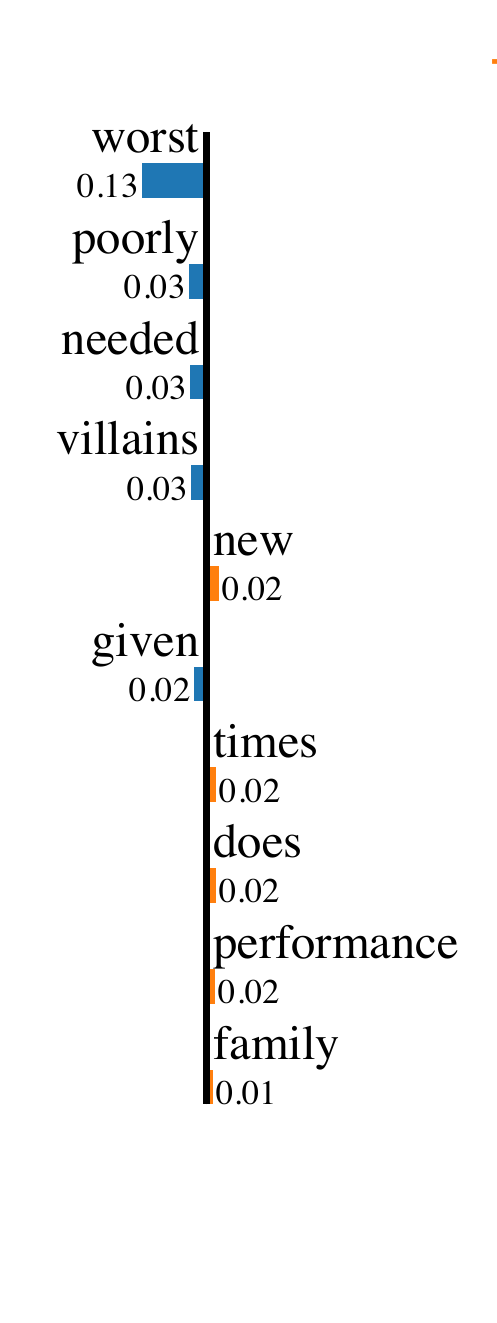}
    \end{center}
    
    \end{subfigure}
    \caption{Comparison of Explanations for TFIDF movie sentiment classifier, GUTEK (left) vs LIME (right) (negative sample id 875)}
    \label{fig:interpretable}
\end{figure*}

While we have illustrated the importance of the granularity hyperparameter in terms of distributional shift, computational complexity and fidelity, the explanations created by different choices of granularity are also inherently different. For example, the sentence-based interpretations give context, while the word-based methods are easier to understand at a glance. In Figure~\ref{fig:interpretable}, we show the interpretations by GUTEK and LIME for a negative movie review given. (TF-IDF based Random-Forest Classifier is used, further examples are given in Appendix~\ref{appendix:qualitative})
We can see that both approaches correctly identify \textit{worst} as a key driver for negative prediction. However, since the sentence also contains the context, giving it as explanation also provides the information that it was in fact the \textit{worst villain} and not the \textit{worst screenplay} or \textit{worst story-line}. A similar effect is observed with \textit{poorly}.
Which interpretation is easier to understand may be domain and application specific. Nonetheless, this effect should also be considered when choosing the granularity of NLP interpretability applications.

%% file: sections/conclusion.tex
In this work, we illustrated limits of current state-of-the-art model-agnostic interpretability methods based on word sampling (\textit{e.g.} LIME, SHAP), prone to out-of-distribution sampling when it comes to complex NLP classifiers like BERT and questioning the truthfulness of the explanations.
Word-based sampling also suffers from high computational complexity, limiting the exploration of the neighborhood of the text whose prediction is to explain.
These limitations are addressed with a sentence-based approach resulting in better fidelity.
The main take-aways thus are (1) the challenges arising with word-based approaches (distributional shift, computational complexity and human interpretability) and (2) the illustration that a simple sentence based model (GUTEK) attains improved performance compared to word-based methods.

%% file: sections/appendix.tex
\section{Reproducibility}
\label{appendix:reproducibility}
To ensure reproducibility, we give the implementation details of our experiments. Direct implementations can also be found directly on our Github\footnote{\url{https://github.com/axa-rev-research/gutek}}.
\subsection{The Case Against Word-Based Black-Box Interpretability}
\subsubsection{Distributional Shift}
We use the last embedding of the classification token as representation of the whole text. We use base uncased BERT~\citep{devlin2019bert}. For the visualisation experiment, we directly use this embedding to calculate Wasserstein distance. To visualize, we use t-SNE on the combined dataset (word removed + sentence removed + original) with PCA initialisation and a perplexity of 100. The algorithm is given a maximum of 5000 iterations, for other parameters we used SKLearn~\citep{scikit-learn} defaults.

For evaluating distributional shift with classifier accuracy, we use base uncased BERT~\citep{devlin2019bert}, base RoBERTa~\citep{liu2019roberta}, base uncased DistilBERT~\citep{sanh2019distilbert} and the small ELECTRA~\citep{clark2020electra} discriminator. The text embeddings are pairwise used to create a classification problem, which uses a random 75-25 train test split. We train a Random Forest Classifier using default SKLearn parameters, controlling for complexity using the maximum depth with options 2, 5, 7, 10, 15 and 20. The best choice is selected using out-of-bag accuracy. Results in Figure~\ref{fig:ood:classifier} and Table~\ref{tab:ood} represents performance on the test-set.

\subsubsection{Computational Complexity}
In order to have normal flowing text, we use text from Wikipedia, notably contexts from SQuAD 2.0~\citep{rajpurkar2018squad}. We compare the number of sentences and the number of words, obtained using NLTK~\citep{bird2009natural} \textit{sent\_tokenize} and \textit{word\_tokenize} respectively.

\subsection{Experiments and Analysis}
\subsubsection{Fidelity Evaluation with QUACKIE}
\label{appendix:reproducibility:gutek}
We use code provided with QUACKIE~\citep{rychener2020} to test GUTEK. In our implementation of GUTEK, we use NLTK \textit{sent\_tokenize} to split the text into sentences and use the SKLearn implementation of the Linear Regression as surrogate. The coefficients of the linear regression are used as sentence scores.


\section{Tabular Results for OOD Classification}
In addition to plotting, we give the results from Figure~\ref{fig:ood:classifier} in Table~\ref{tab:ood}.

\begin{table}[t]
\begin{center}
\begin{small}
\begin{sc}
\begin{tabular}{llrr}
\toprule
dataset &          lm &  word &  sentence \\
\midrule
\multirow{4}{*}{Wikipedia} &        BERT &  0.81 &      0.57 \\
&  DistilBERT &  0.79 &      0.56 \\
&     ROBERTA &  0.86 &      0.70 \\
&     ELECTRA &  0.92 &      0.60 \\
\midrule
\multirow{4}{*}{Amazon}  &        BERT &  0.81 &      0.59 \\
&  DistilBERT &  0.79 &      0.56 \\
&     ROBERTA &  0.87 &      0.76 \\
&     ELECTRA &  0.92 &      0.60 \\
\midrule
\multirow{4}{*}{NYT} &        BERT &  0.81 &      0.59 \\
&  DistilBERT &  0.78 &      0.57 \\
&     ROBERTA &  0.87 &      0.76 \\
&     ELECTRA &  0.91 &      0.61 \\
\midrule
\multirow{4}{*}{Reddit}  &        BERT &  0.79 &      0.59 \\
&  DistilBERT &  0.78 &      0.56 \\
&     ROBERTA &  0.86 &      0.76 \\
&     ELECTRA &  0.92 &      0.60 \\
\bottomrule
\end{tabular}
\end{sc}
\end{small}
\end{center}
\caption{OOD Classification Accuracy in Tabular Form}
\label{tab:ood}
\vskip -0.1in
\end{table}

\section{Qualitative Evaluation}
\label{appendix:qualitative}
\begin{figure*}
    \begin{subfigure}[c]{0.6\linewidth}
    \begin{center}
    \includegraphics[width=\linewidth, trim=5cm 0 24cm 0, clip]{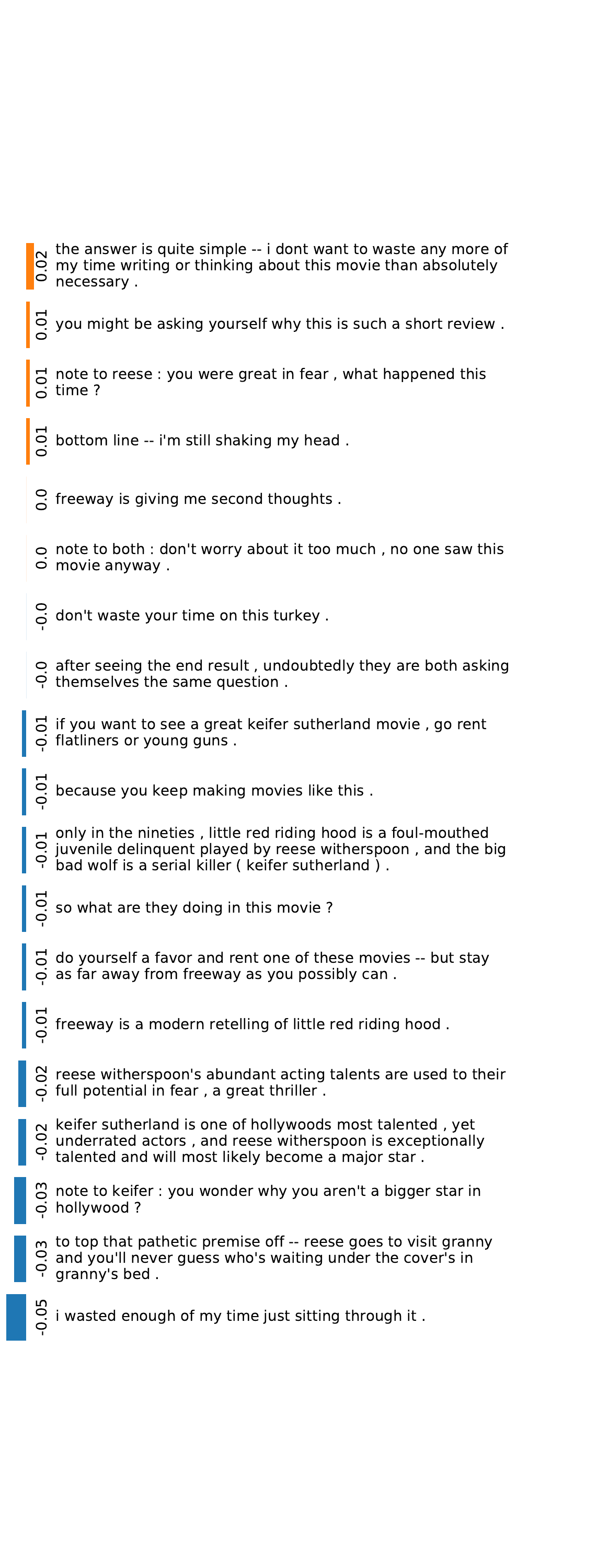}
    \end{center}
    \end{subfigure} 
    \begin{subfigure}[c]{0.3\linewidth}
    \begin{center}
        \includegraphics[width=\linewidth, trim=13cm 20cm 49cm 1cm, clip]{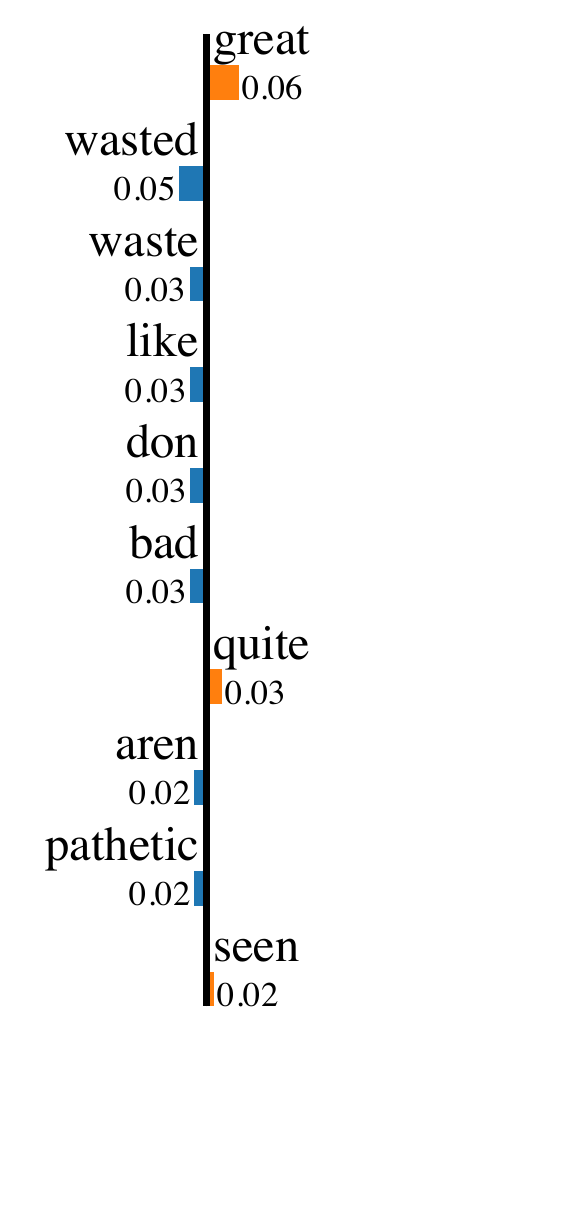}
    \end{center}
    
    \end{subfigure}
    \caption{Comparison of Explanations for TFIDF movie sentiment classifier, GUTEK (left) vs LIME (right) (sample id 370)}
    \label{fig:interpretable:a2}
\end{figure*}

\begin{figure*}
    \begin{subfigure}[c]{0.6\linewidth}
    \begin{center}
    \includegraphics[width=\linewidth, trim=5cm 4cm 24cm 4cm, clip]{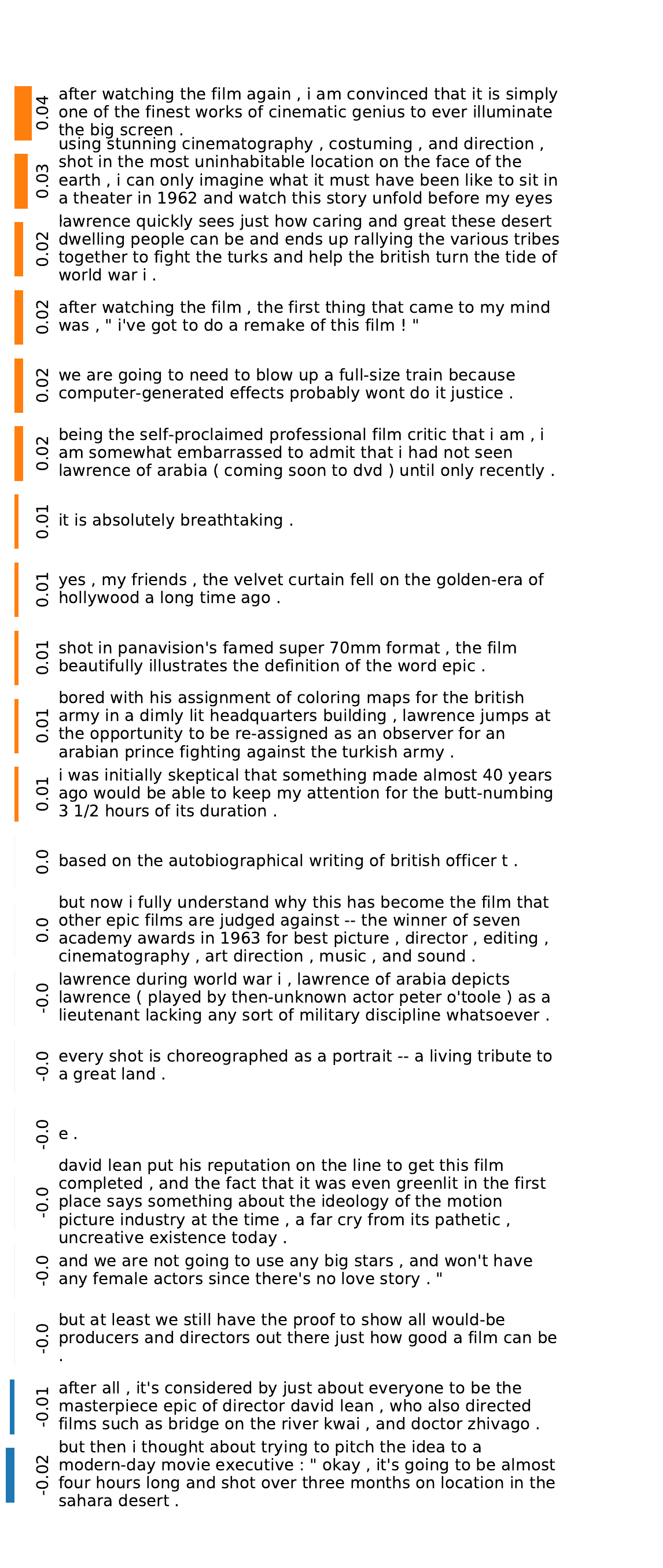}
    \end{center}
    \end{subfigure} 
    \begin{subfigure}[c]{0.3\linewidth}
    \begin{center}
        \includegraphics[width=\linewidth, trim=13cm 20cm 49cm 1cm, clip]{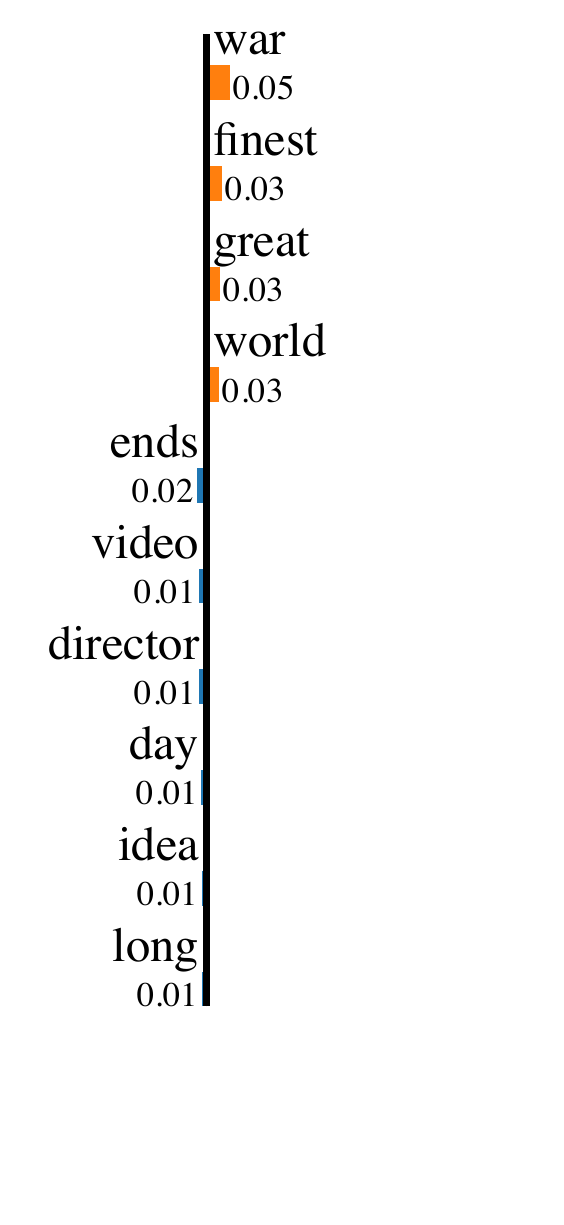}
    \end{center}
    
    \end{subfigure}
    \caption{Comparison of Explanations for TFIDF movie sentiment classifier, GUTEK (left) vs LIME (right) (sample id 70)}
    \label{fig:interpretable:a3}
\end{figure*}

In Figures~\ref{fig:interpretable:a2} and~\ref{fig:interpretable:a3}, we give some more illustrations of the different explanations, similarly to Figure~\ref{fig:interpretable}

\section{Complete QUACKIE results}
\label{appendix:quackie results}
We also give results for all datasets in QUACKIE and report the scores for all other methods currently in QUACKIE in Tables~\ref{tab:res:iou}, \ref{tab:res:hpd} and \ref{tab:res:snr}.

\clearpage

\begin{table*}[t]
    \centering
    \begin{adjustbox}{width=\textwidth}
    \begin{tabular}{lll|rr|rr|rr|rr|rr}
        &&&\multicolumn{2}{c|}{SQuAD}&\multicolumn{2}{c|}{New Wiki}&\multicolumn{2}{c|}{NYT}&\multicolumn{2}{c|}{Reddit}&\multicolumn{2}{c}{Amazon}\\
        Interpreter & Aggregation & Samples & Classif & QA& Classif & QA& Classif & QA& Classif & QA& Classif & QA\\
        \hline
        \hline
        \multirow{2}*{\textbf{GUTEK}}& \multirow{2}*{-} & 10 &88.55&90.53&87.7&89.54&87.66&88.04&71.62&76.86&79.09&78.95\\
        && 100 &91.38&90.53&90.83&90.53&91.56&91.84&80.62&86.98&84.77&86.14\\
        \hline
        \hline
        \multirow{4}*{LIME\textsuperscript{\dag}} & \multirow{2}*{sum} & 10 & 37.70& 38.47 &40.72&41.40&40.22&41.98&31.35&35.34&32.69&35.27\\
                            &                    & 100 & 58.04 & 69.90&60.74&70.82&62.02&73.50&50.33&69.02&53.57&67.58\\
                            & \multirow{2}*{max} & 10 &34.06&35.36&36.98&37.77&36.19&36.43&26.52&28.89&29.80&32.12\\
                            &                    & 100 &57.86&68.30&59.65&68.43&61.57&72.00&48.22&65.23&53.09&66.24\\
        \hline
        \multirow{4}*{SHAP\textsuperscript{\dag}} & \multirow{2}*{sum} & 10 & 30.48 & 32.90&31.66&32.84&29.26&31.03&22.13&23.75&24.59&25.43\\
                            &                    & 100 & 54.85 & 65.92&57.53&65.79&56.38&67.70&49.35&65.02&54.03&67.68\\
                            & \multirow{2}*{max} & 10 &  29.69&30.81&30.72&31.68&28.32&30.00&21.17&22.58&22.72&23.84\\
                            &                    & 100 &52.45&62.34&54.56&63.18&53.19&64.78&45.79&60.03&49.54&63.35\\
        \hline
        \hline
        \multirow{2}*{Saliency} & sum & - & 74.74 & 91.12 &72.19&91.18&68.87&88.46&57.57&85.26&64.82&85.91\\
                                & max & - &66.27&80.79&65.04&80.78&58.95&76.07&48.41&77.33&59.52&79.70\\
        \hline
        \multirow{2}*{Integrated Gradients} & sum & 50 & 66.73& 85.93& 65.00& 85.93& 65.44& 85.20& 51.62& 79.73& 51.96& 78.21\\
                                            & max & 50 &62.73& 87.05& 60.70& 86.73& 61.63& 85.92& 50.24& 82.35& 49.09& 82.45\\
        \hline
        \multirow{2}*{SmoothGrad} & sum & 5 & 60.98&91.28 &60.29&90.56&60.25&88.29&50.32&84.51&52.34&84.40\\
                                  & max & 5 &59.48&82.16&61.45&82.38&56.93&78.33&45.95&77.72&53.03&78.26\\
        \hline
        \hline
        Random & - & - & 24.64&25.38 &26.86&27.39&24.53&24.36&16.51&16.09&18.71&19.17\\
    \end{tabular}
    \end{adjustbox}
    \caption{IoU Results
    }
    \label{tab:res:iou}
\end{table*}

\begin{table*}[t]
    \centering
    \begin{adjustbox}{width=\textwidth}
    \begin{tabular}{lll|rr|rr|rr|rr|rr}
        &&&\multicolumn{2}{c|}{SQuAD}&\multicolumn{2}{c|}{New Wiki}&\multicolumn{2}{c|}{NYT}&\multicolumn{2}{c|}{Reddit}&\multicolumn{2}{c}{Amazon}\\
        Interpreter & Aggregation & Samples & Classif & QA& Classif & QA& Classif & QA& Classif & QA& Classif & QA\\
        \hline
        \hline
        \multirow{2}*{\textbf{GUTEK}}& \multirow{2}*{-} & 10 &90.75&92.37&90.17&91.68&89.64&90.02&74.93&79.5&81.76&81.71\\
        && 100 &93.02&92.37&92.72&92.37&93.02&93.36&83.04&88.66&86.84&88.17\\
        \hline
        \hline
        \multirow{4}*{LIME\textsuperscript{\dag}} & \multirow{2}*{sum} & 10 &50.29&50.83&53.32&53.76&51.62&53.12&39.99&43.56&42.31&44.64\\
                            &                    & 100 &66.50&75.98&68.93&76.93&69.17&78.58&56.60&72.97&60.12&72.25\\
                            & \multirow{2}*{max} & 10 &45.12&46.19&47.85&48.62&46.60&47.11&34.47&36.74&38.43&40.63\\
                            &                    & 100 &63.74&71.33&65.47&71.89&67.28&75.25&53.23&67.65&58.41&69.21\\
        \hline
        \multirow{4}*{SHAP\textsuperscript{\dag}} & \multirow{2}*{sum} & 10 &41.22&44.09&42.87&44.57&39.06&41.38&28.94&31.26&32.97&34.63\\
                            &                    & 100 &63.93&72.75&66.18&72.91&64.39&73.74&55.59&69.44&60.48&72.29\\
                            & \multirow{2}*{max} & 10 &37.74&39.28&39.29&40.68&36.40&38.54&27.30&29.13&30.24&32.05\\
                            &                    & 100 &59.85&67.47&61.80&68.35&60.80&69.98&51.19&63.41&55.32&66.86\\
        \hline
        \hline
        \multirow{2}*{Saliency} & sum & - &79.91&93.01&78.20&93.06&74.97&90.71&63.05&87.21&69.96&88.01\\
                                & max & - &72.99&84.83&72.40&84.81&66.86&80.94&55.10&80.32&65.31&82.74\\
        \hline
        \multirow{2}*{Integrated Gradients} & sum & 50 &73.52& 88.85& 72.39& 88.93& 71.99& 88.00& 57.84& 82.46& 58.93& 81.56\\
                                            & max & 50 &70.15& 89.67& 68.93& 89.48& 68.73& 88.52& 56.51& 84.70& 56.35& 85.10\\
        \hline
        \multirow{2}*{SmoothGrad} & sum & 5 &69.03&93.08&68.92&92.59&68.05&90.52&56.77&86.58&59.36&86.72\\
                                  & max & 5 &67.83&85.77&69.64&86.12&65.32&82.65&53.00&80.64&59.84&81.43\\
        \hline
        \hline
        Random & - & - &40.28&40.71&42.66&42.92&39.23&39.18&27.41&27.17&30.79&31.34\\
    \end{tabular}
    \end{adjustbox}
    \caption{HPD Results
    }
    \label{tab:res:hpd}
\end{table*}

\begin{table*}[t]
    \centering
    \begin{adjustbox}{width=\textwidth}
    \begin{tabular}{lll|rr|rr|rr|rr|rr}
        &&&\multicolumn{2}{c|}{SQuAD}&\multicolumn{2}{c|}{New Wiki}&\multicolumn{2}{c|}{NYT}&\multicolumn{2}{c|}{Reddit}&\multicolumn{2}{c}{Amazon}\\
        Interpreter & Aggregation & Samples & Classif & QA& Classif & QA& Classif & QA& Classif & QA& Classif & QA\\
        \hline
        \hline
        \multirow{2}*{\textbf{GUTEK}}& \multirow{2}*{-} & 10 &39.48&37.37&42.63&37.86&32.21&30.68&19.12&17.69&26.11&22.22\\
        && 100 &35.49&37.37&39.5&37.37&30.38&33.13&18.22&19.48&20.9&22.92\\
        \hline
        \hline
        \multirow{4}*{LIME} & \multirow{2}*{sum} & 10 &39.23&38.20&41.82&40.66&36.98&37.26&25.89&22.84&27.87&27.08\\
                            &                    & 100 &39.30&40.91&42.30&43.96&39.41&39.38&32.90&46.71&27.42&32.52\\
                            & \multirow{2}*{max} & 10 &91.76&91.54&94.38&88.83&93.24&85.01&107.89&110.55&93.77&95.83\\
                            &                    & 100 &125.98&176.07&124.96&162.51&133.54&184.66&171.42&305.21&151.52&232.94\\
        \hline
        \multirow{4}*{SHAP} & \multirow{2}*{sum} & 10 &73.24&67.24&74.17&67.85&71.42&68.34&91.28&83.95&68.02&60.68\\
                            &                    & 100 &42.27&42.09&44.80&45.60&37.31&43.69&34.30&40.05&28.76&34.64\\
                            & \multirow{2}*{max} & 10 &99.16&102.31&97.42&101.44&97.10&92.66&130.98&127.01&99.87&102.16\\
                            &                    & 100 &107.51&137.77&102.01&132.57&94.43&132.64&149.95&240.47&135.62&207.62\\
        \hline
        \hline
        \multirow{2}*{Saliency} & sum & - &37.29&39.92&40.88&40.14&35.14&34.23&19.10&19.77&22.90&23.34\\
                                & max & - &35.58&38.20&39.75&39.81&34.08&36.35&19.15&20.04&22.17&23.78\\
        \hline
        \multirow{2}*{Integrated Gradients} & sum & 50 &37.28& 37.32& 39.16& 40.19& 33.30& 33.04& 18.96& 20.60& 22.50& 24.60\\
                                            & max & 50 &35.71& 34.99& 38.74& 38.09& 32.55& 32.80& 18.41& 20.57& 21.80& 23.69\\
        \hline
        \multirow{2}*{SmoothGrad} & sum & 5 &38.13&37.22&41.29&40.15&35.16&33.04&19.40&20.33&23.34&23.11\\
                                  & max & 5 &37.55&36.61&40.29&40.04&34.85&35.98&19.23&19.62&22.45&22.69\\
        \hline
        \hline
        Random & - & - &37.70&37.34&40.63&40.52&34.87&35.06&19.24&19.79&23.21&23.70\\
    \end{tabular}
    \end{adjustbox}
    \caption{SNR Results (Examples for which noise cannot be estimated are omitted)}
    \label{tab:res:snr}
\end{table*}